 \documentclass[plmr, twocolumn]{jmlr} 

\usepackage{booktabs}
\usepackage[load-configurations=version-1]{siunitx} 

\usepackage{multirow}

\theorembodyfont{\upshape}
\theoremheaderfont{\scshape}
\theorempostheader{:}
\theoremsep{\newline}

\title[OOD Detection for Medical Applications: Guidelines for Practical Evaluation]{Out-of-Distribution Detection for Medical Applications: Guidelines for Practical Evaluation}
\author{%
\Name{Karina Zadorozhny} \Email{karina.zadorozhny@gmail.com}\\
\addr Pacmed BV - Amsterdam, The Netherlands
\AND
\Name{Patrick Thoral} \Email{p.thoral@amsterdamumc.nl}\\ 
\Name{Paul Elbers} \Email{p.elbers@amsterdamumc.nl}\\
\addr Department of Intensive Care Medicine,              Laboratory for Critical Care Computational Intelligence              (LCCCI), Amsterdam Medical Data Science (AMDS), Amsterdam              UMC, Vrije Universiteit, Amsterdam, The Netherlands\\
\Name{Giovanni Cinà} \Email{giovanni.cina@pacmed.nl}\\
\addr Pacmed BV - Amsterdam, The Netherlands
}

\begin{document}

\maketitle

\begin{abstract}

    Detection of Out-of-Distribution (OOD) samples in real time is a crucial safety check for deployment of machine learning models in the medical field. Despite a growing number of uncertainty quantification techniques, there is a lack of evaluation guidelines on how to select OOD detection methods in practice. This gap impedes implementation of OOD detection methods for real-world applications.
    Here, we propose a series of practical considerations and tests to choose the best OOD detector for a specific medical dataset. These guidelines are illustrated on a real-life use case of Electronic Health Records (EHR). Our results can serve as a guide for implementation of OOD detection methods in clinical practice, mitigating risks associated with the use of machine learning models in healthcare.
\end{abstract}


\section{Introduction}
\label{sec:intro}

    When deploying a machine learning model in a high-risk environment such as healthcare, it is crucial to reduce the risks of giving an incorrect prediction. One of the most common causes of failures of a well-performing model is dissimilarity between training (in-distribution) data and data that the model is used on after deployment \citep{saria2019tutorial}. This problem is further exacerbated in a non-stationary medical environment: the performance of predictive models can drop drastically when evaluated on medical data collected just a couple of years removed from in-distribution data \citep{nestor2018rethinking}. The reason for this can be, for example, a change in the demographics of the target population, changes in treatment protocols, or changes in data collection procedures \citep{finlayson_clinician_2021}. These examples constitute a covariate shift where feature distributions changed compared to in-distribution data \citep{shimodaira2000improving, moreno2012unifying}. For reliable use of prediction models in healthcare it is necessary to have the ability to detect these changes in real-time.

    In the past five years, there has been a surge in methods for OOD detection and uncertainty quantification. Despite the progress, OOD detection methods often fail at flagging atypical inputs. While most studies have focus on image data \citep{ovadia_can_2019, nalisnick2018deep}, it has recently been shown that many OOD detection methods fail to distinguish OOD inputs on medical tabular data \citep{ulmer2020trust}. 
    
    
    While predictive models are tailored and tested on specific datasets, there is no universal way to evaluate OOD detection methods in practice. This gap impedes wide-spread implementation of OOD detection methods. 
    In this paper, we provide a practical set of guidelines on how to select an OOD detector in a given medical AI application. Our specific contributions are the following:

    \begin{itemize}
        \item We describe how to take into account dataset-specific variables when evaluating OOD detector models in practice.
        \item We show how to test OOD detectors on distinct families of OOD groups that can be created from available data using inclusion-exclusion criteria and withholding  samples during training.
        \item We describe how to to use interpretability tools to assess whether influential features that drive novelty predictions are clinically relevant.
        \item We provide an open-source repository that can be used to evaluate OOD detection methods for any mixed-type tabular dataset.\footnote{\tiny  \href{https://github.com/Giovannicina/selecting_OOD_detector}{https://github.com/Giovannicina/selecting\_OOD\_detector}}
    \end{itemize}

    We illustrate these general principles on real-world EHR data from the Intensive Care Unit (ICU) of Amsterdam University Medical Centers (AmsterdamUMC) \citep{thoral_sharing_2021}. We show that using the guidelines proposed in this paper,  we can detect underperforming OOD detectors and select the best methods to be deployed in practice.


\section{Related Work}
\label{sec:related_work}
        
    Despite the impressive performance of machine learning models, recent studies have highlighted the potential risks of undetected OOD data points in critical areas such as healthcare. Besides deliberate threats such as data poisoning and adversarial attacks \citep{papangelou_toward_2019, finlayson_adversarial_2019, mozaffari-kermani_systematic_2015}, deployed models can suffer from severe malfunction due to dataset shifts. \citet{finlayson_clinician_2021} described different scenarios that lead to dataset shifts in practice and highlighted the need for more robust and reliable models in healthcare.
    
    To mitigate risks of failure due to dataset shift caused by covariate shifts \citep{bickel_discriminative_2009}, recent studies proposed a plethora of uncertainty quantification methods. These methods can be broadly divided into two groups based on the type of uncertainty they express. The first group uses predictive uncertainty (\textit{uncertainty about predictions}) which is model-specific and indicates how confident a model is in its prediction \citep{ovadia_can_2019}; such uncertainty can be measured with a variety of metrics. Models in this category include Bayesian Neural Networks \citep{blundell2015weight}, Ensemble Neural Networks \citep{lakshminarayanan2017simple}, or  Monte Carlo Dropout \citep{gal2016dropout}.
    The second group of models, density estimators, expresses \textit{uncertainty about samples} by learning the distribution of training data. Models such as Variational Autoencoder (VAE; \citeauthor{kingma2014autoencoding},  \citeyear{kingma2014autoencoding};
    \citeauthor{ran2020detecting}, 
    \citeyear{ran2020detecting}) or Normalizing Flows \citep{pmlr-v37-rezende15} can be used to flag samples with low likelihood under an estimated in-distribution function. 
    

    There has been only limited work published to date that focuses on the practical evaluation of OOD detection methods in real-world scenarios. \cite{hendrycks2020scaling} proposed a benchmark for OOD detection in the imaging domain that could replace small-scale datasets containing only a few classes with a more realistic high-scale multi-class anomaly segmentation task. \cite{techapanurak2021practical} described practical tests of OOD detection in the imaging domain to aid the evaluation of models in real-world applications. However, no guidelines on how to test OOD detectors are available for medical tabular data. In this paper, we aim to close this gap between research studies and the practical implementation of OOD detection methods in a medical setting.

\section{Methods}
\label{sec:methods}
    We first describe considerations that should be taken into account when choosing an OOD detector for a specific application. In Section \ref{sec:methods:ood_tests}, we put forward  guidelines for creating tests to assess the performance of OOD detection methods. 
    
    The proposed OOD tests are illustrated on a real-world use case of EHR data of AmsterdamUMC. Details about the dataset, experimental set-up, implemented models, and the way OOD groups were selected for this dataset are described in Appendix \ref{apd:methods}.
    
        \begin{table*}
        \scalebox{0.83}{
        \centering
        \begin{tabular}{|lllll|} 
        \hline
        \begin{tabular}[c]{@{}l@{}}\textbf{Type of }\\\textbf{~OOD groups}\end{tabular}    &   & \textbf{Examples}  & \textbf{How To Create?} & \begin{tabular}[c]{@{}l@{}}\textbf{AmsterdamUMC }\\\textbf{Example}\end{tabular} \\ 
        \hline
        \multirow{2}{*}{\begin{tabular}[c]{@{}l@{}}Outside\\inclusion-\\exclusion\\criteria\end{tabular}} & Static      & Age, race                 & \begin{tabular}[c]{@{}l@{}}Separate excluded groups\\according to demographic-\\related characteristics.~\end{tabular} & \begin{tabular}[c]{@{}l@{}}\end{tabular}    \\  & Dynamic    & \begin{tabular}[c]{@{}l@{}}Treatment,\\ length of stay\end{tabular}                              & \begin{tabular}[c]{@{}l@{}}Split on one or more features~\\related to a clinical status that \\is subject to change.\end{tabular}                                                                                              & \begin{tabular}[c]{@{}l@{}}Ventilation,\\medications,\\patients far\\ from discharge.\\\end{tabular}  \\ 
        \hline
        New disease     &      & \begin{tabular}[c]{@{}l@{}}New infectious\\disease, any diagnosis\\underrepresented in\\training data\\\end{tabular} & \begin{tabular}[c]{@{}l@{}} Artificially create \\an OOD group by withholding\\them during training.\end{tabular}                                & \begin{tabular}[c]{@{}l@{}}COVID-19 patients \\and  suspects.\end{tabular}      \\ 
        \hline
        \end{tabular}
        }
        \caption{Clinically relevant OOD groups.}
        \label{table:methods_ood_tests}
        \end{table*}
        
    \subsection{Considerations}
        \label{sec:methods:recs}
        
        \subsubsection{Model Type}

        It has been shown that models that utilize predictive uncertainty in a classification setting can be overconfident in their faulty predictions \citep{guo2017_calibration, ovadia_can_2019}. Specifically, a group of discriminator models  assign high predictive confidence to regions with no training data \citep{hein2019relu, ulmer2021know}. Given these findings, we advise using density estimators instead of models that express predictive uncertainty. While there is a growing interest in the limitations of density estimators for OOD detection \citep{nalisnick2018deep,kirichenko_why_2020, zhang_understanding_2021}, density models outperformed discriminator models on several OOD detection tasks on tabular medical data \citep{ulmer2020trust}.

        \subsubsection{Data Type}
        Many density estimators can perform well on continuous data but are not directly suitable for discrete distributions. Research efforts of extending models to discrete distributions most commonly focus on text inputs \citep{tran_discrete_2019, miao_neural_2016}. For some applications, such as learning density of images, dequantization can be applied to map discrete pixel values data to continuous distributions \citep{hoogeboom_learning_2020}. However, this approach is not applicable to categorical data with no intrinsic order.
        
        Mixed type data that consist of categorical and continuous features pose additional challenges to density modeling \citep{nazabal_handling_2020}. Therefore, the proportion of categorical features should be taken into account when choosing an OOD detector.

        \subsubsection{Dimensionality and Size of Data}
        The dimensionality of data is a crucial factor that influences the performance of density estimators. For example,  standard kernel density estimators can achieve very good results on low-dimensional inputs but their performance degrades exponentially with an increased number of dimensions \citep{nagler_evading_2016, wang_nonparametric_2019}. Other models, such as Autoencoder (AE) and Probabilistic Principal Component Analysis (PPCA) that perform dimensionality reduction by default are less affected by the number of features.
        
        Having a large number of samples from which to estimate data distribution can unequally affect different density estimators. While most models benefit from having access to more data, algorithms such as Local Outlier Factor (LOF) that measure local density of samples using k-nearest-neighbors algorithm become very time- and memory- inefficient with an increasing number of samples \citep{de_vries_finding_2010}. 

        \subsubsection{Class Balance}
        In the classification setting, having unbalanced classes, which is very common for medical diagnostics, can negatively influence usefulness of learned data density. If there is not a sufficient number of samples of one class present in the training dataset, new samples of this minority class can be mistakenly flagged as OOD  despite being an essential component of the dataset. While densities of two classes could be modeled separately or using a class-conditional density estimator \citep{learning_sohn_2015}, this approach is not feasible as
        we would have to first predict a sample's class and only then test whether the sample is in-distribution. Therefore, likelihood scores for underrepresented classes should be monitored when comparing different density models.
        

    \subsection{Designing OOD Tests}
    \label{sec:methods:ood_tests}
    
    \begin{figure*}[ht!]
            \centering
            \includegraphics[width=2\columnwidth]{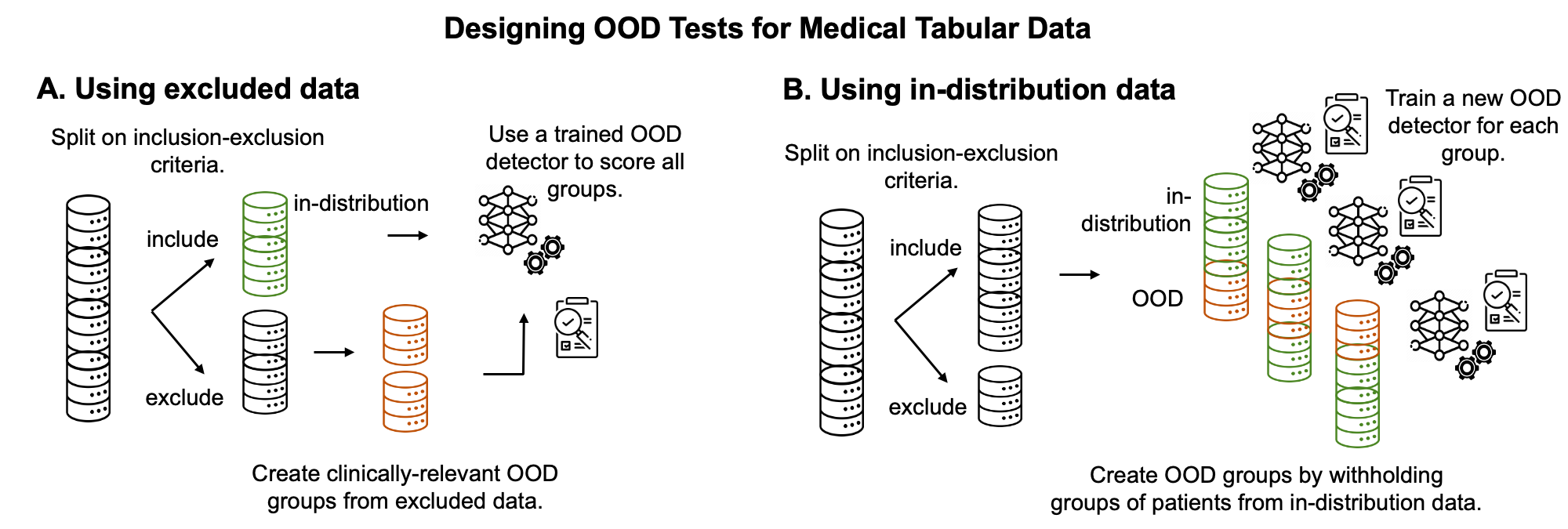}
            \caption{OOD groups can be created from data that was excluded under inclusion-exclusion criteria (panel A). If there are not enough samples in excluded data, groups can be withheld during training (panel B).
            }
            \label{fig:ood_design}
        \end{figure*}
        While any data point outside the training distribution is considered as OOD, it is not feasible to test against \textit{all} possible inputs. Work by \citet{zhang_understanding_2021} argues that it is not theoretically possible to design a method that could detect all possible out-distribution inputs.  
        \cite{winkens2020contrastive} distinguished \textit{far}-OOD and \textit{near}-OOD depending on how similar inputs are to in-distribution data. Given that the latter group is more relevant and difficult to detect, we focus on \textit{near}-OOD scenarios, specifying families of OOD groups that could be encountered in a medical setting in real life (Appendix Figure \ref{fig:scheme}). 
    
        We suggest to create  clinically relevant OOD tests  to cover the following categories (Table \ref{table:methods_ood_tests}):
        
        \begin{itemize}
            \item Detecting patients outside static inclusion-exclusion criteria (e.g. demographics).
            \item Detecting patients outside dynamic inclusion-exclusion criteria (e.g. a treatment protocol).
            \item Detecting patients with a new or underrepresented disease.
        \end{itemize}
        
        The OOD groups can be created either by using data eliminated under inclusion-exclusion principle or by withholding specific groups from in-distribution (Figure \ref{fig:ood_design}).
    
        \subsubsection{Patients Outside Static Inclusion-Exclusion Criteria}
        Processing medical data involves defining inclusion and exclusion criteria that specify which samples are going to be part of the cohort to be analyzed. Inclusion criteria, which are typically clinical or demographic characteristics, must be present in order to include a sample. Exclusion criteria eliminate samples that passed inclusion criteria but have other characteristics that could invalidate downstream analysis \citep{patino_inclusion_2018}. 
        
        Inclusion-exclusion criteria can be used to evaluate the performance of OOD detection methods. Provided that models are trained only on samples that follow inclusion-exclusion criteria, we can treat ineligible samples as OOD and compare how well these groups are flagged by OOD detectors. Such groups can be made using demographic or clinical characteristics (Figure \ref{fig:ood_design}A).
    
        \subsubsection{Patients Outside Dynamic Inclusion-Exclusion Criteria}
        \label{sec:methods:ood-test:dyn}
        Another example of a clinically relevant OOD group is patients that should currently be excluded from data, but this status can change over time. In a medical setting, this is a common occurrence and can be caused by a change in the treatment protocol. 

        If data about individual patients are available over a period of time, such groups can be created by finding a criterion that is not static and is subject to change. Alternatively, OOD groups can be created by splitting data on a received treatment. 

\begin{figure*}[ht!]
            \centering
            \includegraphics[width=2\columnwidth]{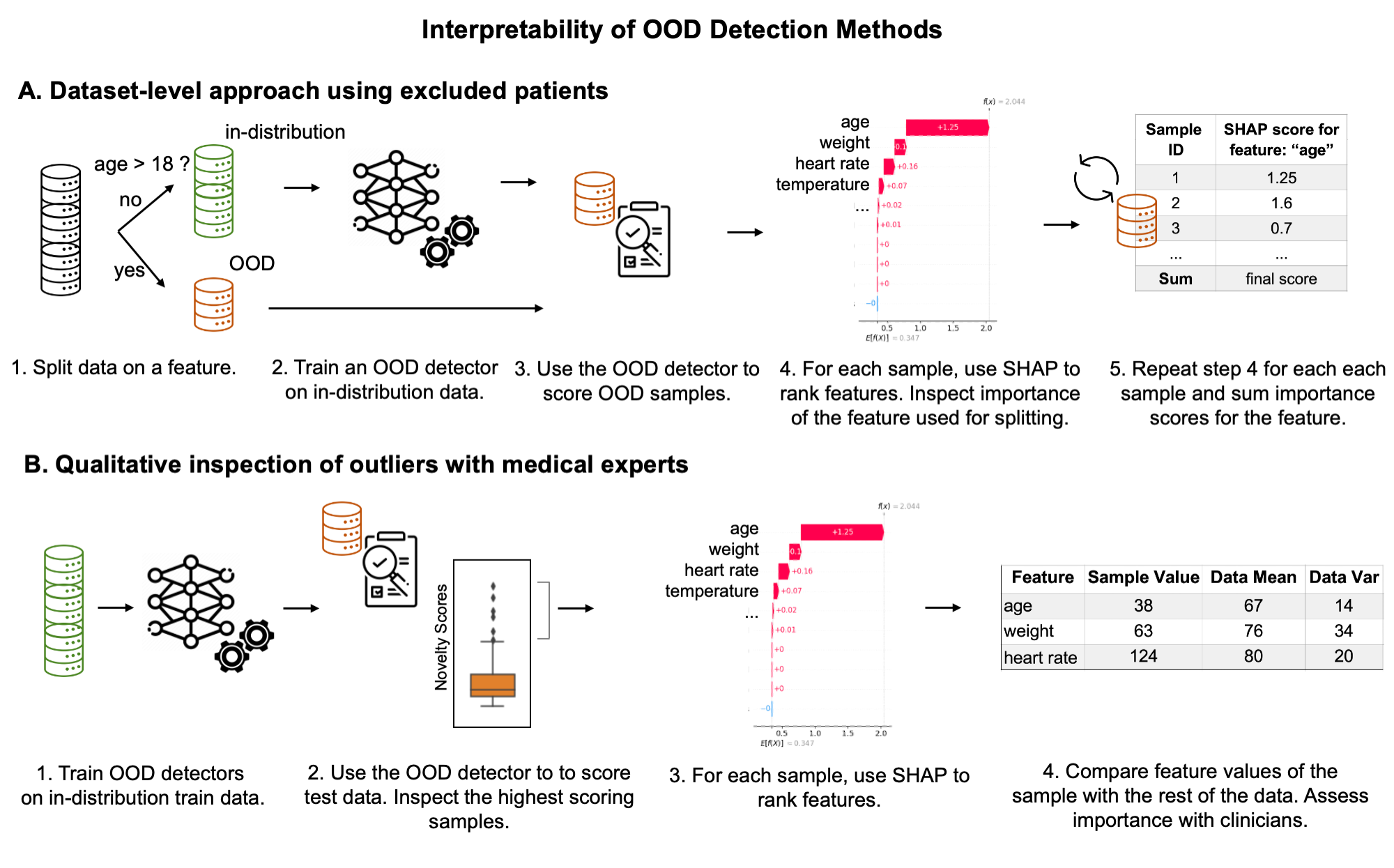}
            \caption{Assessing interpretability of OOD detectors. See Section \ref{methods:ood_test:interp} for description.}
            \label{fig:interp}
        \end{figure*}
        \subsubsection{Patients with New Diseases}
        A clinically relevant scenario that highlights the necessity of OOD detection methods is an occurrence of novel diseases. COVID-19 provides an example of the importance of being able to detect atypical symptoms in real-time instead of giving a prediction and only assessing data retrospectively. A common scenario occurs when a model is used for patients with a disease that was not sufficiently represented in the in-distribution data. 
        We advise creating an OOD group of patients with underrepresented disease by either applying inclusion-exclusion criteria (Figure \ref{fig:ood_design}A) or withholding samples of patients with a specific disease from in-distribution training data (Figure \ref{fig:ood_design}B).
        
        \subsubsection{Interpretability of OOD Detection Methods}
        \label{methods:ood_test:interp}
        Another consideration when deploying OOD detection methods is how transparent novelty predictions are. In high-stakes areas such as healthcare, it is not only useful to be able to flag a sample as OOD but also to see what set of features makes it different from in-distribution data.
        
        Some density estimators are inherently interpretable. For example, for AE-based models, it is possible to inspect features that received the highest reconstruction error. To provide a comparative evaluation of interpretability for all models, we suggest using Shapley Additive Explanations  (SHAP; \citeauthor{lundberg2017unified}, \citeyear{lundberg2017unified}). 
        
        We describe two approaches of assessing interpretability of OOD detection methods (Figure \ref{fig:interp}). The first, dataset-level approach, splits data on one feature to divide in-distribution and OOD data. After scoring OOD samples, SHAP is used to determine feature importance for the assigned novelty scores. Models are then scored on the importance of the feature that was used for splitting (Figure \ref{fig:interp}A).

        In the second, qualitative approach, OOD detectors are trained on in-distribution data and are used to score test data (Figure \ref{fig:interp}B). Samples that receive the highest novelty scores (outliers) are then inspected individually. 
        Ranking the most influential features using SHAP and having medical professionals compare values of these features can help validate whether models are influenced by clinically relevant features and verify that outliers are indeed meant to be in-distribution.

         \subsubsection{Inference Time for Real-Time Prediction}
          The ability to flag OOD samples in real-time, as compared to detecting batches of atypical samples retrospectively, is an important consideration for the deployment of OOD detection methods. While inference time for individual predictions tends not be a limiting factor, when coupled with explainability tools such as SHAP, can significantly slow down the prediction process. 
    \subsection{Experimental setup}
   To illustrate the above-described principles in practice, we put them to test on the AmsterdamUMC dataset. The downstream prediction task is a binary classification of readmission  to ICU after discharge or mortality. For this task, any predictive model can be used while we aim to compare different OOD detection on this dataset (see Appendix \ref{apd:methods} for more details about the dataset and the experimental set-up).
   
   The density estimators compared in the following experiments include AE, VAE \citep{kingma2014autoencoding}, Deterministic Uncertainty Estimator (DUE;  \citeauthor{vanamersfoort2021improving}, \citeyear{vanamersfoort2021improving}) based on Spectral Normalized Gaussian Process \citep{liu2020simple}, Masked Autoregressive Flow (Flow \citeauthor{papamakarios2018masked}, \citeyear{papamakarios2018masked}), PPCA \citep{bishop_1999_ppca}, and LOF \cite{de_vries_finding_2010}. Description of the models and hyperparameters can be found in Appendix \ref{apd:methods:models}.
    
\section{Results}
\label{sec:results}
        
    
    In the following section we describe the results obtained applying the aforementioned methodology on the AmsterdamUMC data.

\begin{figure}[!t]
            \centering
            \includegraphics[width=\columnwidth]{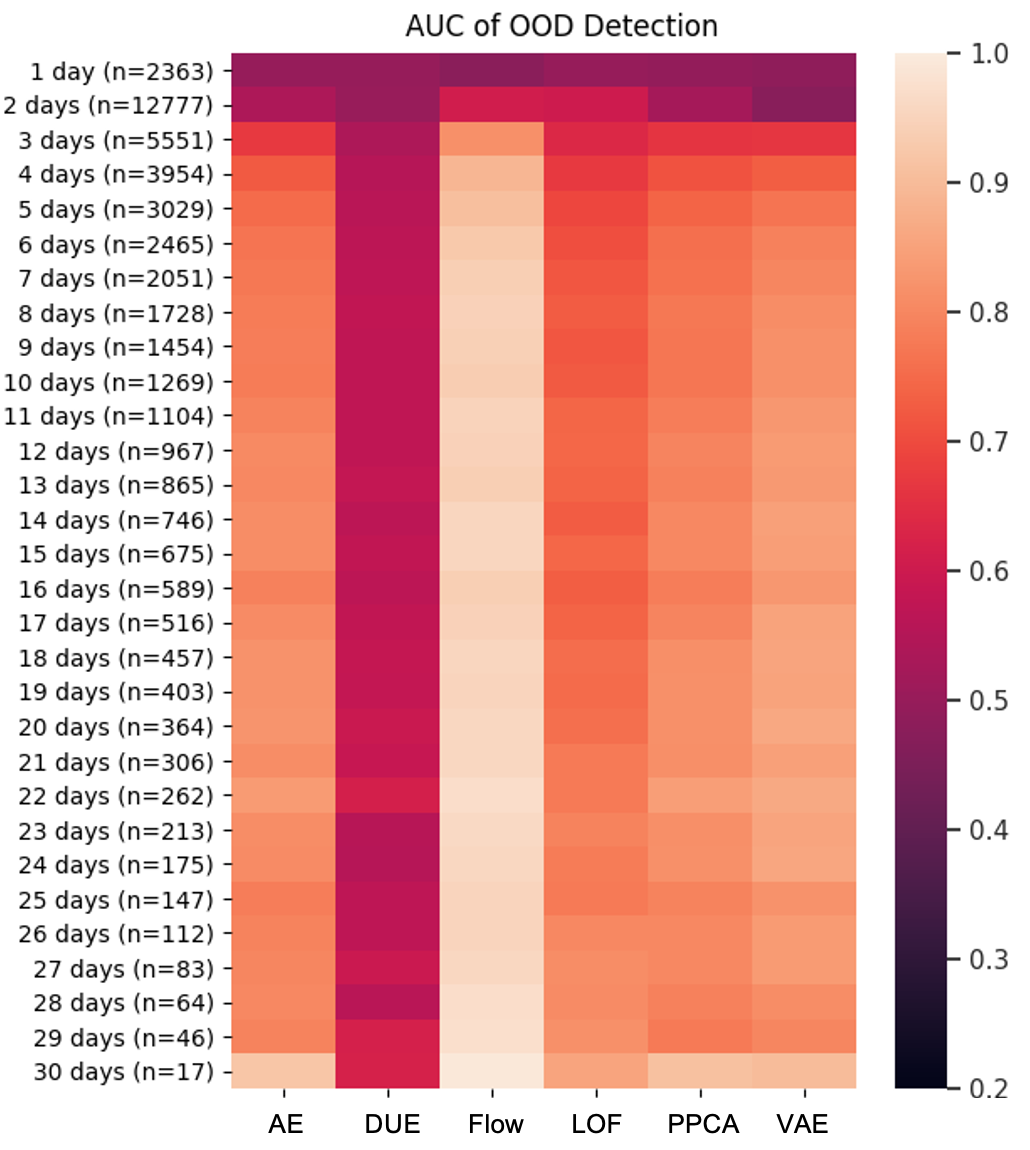}
            \caption{Mean AUC scores ($n$=5) of detecting patients far from discharge. Row labels indicate the number of days before discharge.}
            \label{fig:days_before_discharge}
        \end{figure}

        \textbf{Dynamic Inclusion-Exclusion Criteria: Length of Stay}. 
        We followed the guidelines described in Section \ref{sec:methods:ood-test:dyn} and selected patients outside dynamic inclusion-exclusion criteria which are subject to change in time. We used the fact that the AmsterdamUMC dataset contains retrospective time series data for each patients. The in-distribution data for this experiment are patients in the last day of their ICU admission and OOD groups are separated based on how far patients are from being discharged (Figure \ref{fig:days_before_discharge}). The further patients are from discharge, the more dissimilar their features are compared to in-distribution data. While all models except DUE showed the desired gradient of increasing AUC-ROC score, Flow achieved a near-perfect score already for patients 4-5 days before discharge.
        
        \textbf{Dynamic Inclusion-Exclusion Criteria: Discharge Destination}. 
        Given that the main predictive task is classification of readmission probability, patients that died at the ICU or were transferred to another department or a different hospital are excluded from in-distribution data. 
        
        Most models detected patients with discharge destination of mortuary more easily than patients that were transferred to an ICU of a different hospital (Appendix Figure \ref{fig:appendix_discharge_destin}). This result is reassuring as, arguably, features of transferred patients more closely resemble in-distribution data.

        \begin{figure*}[!t]
            \centering
            \includegraphics[width=2\columnwidth]{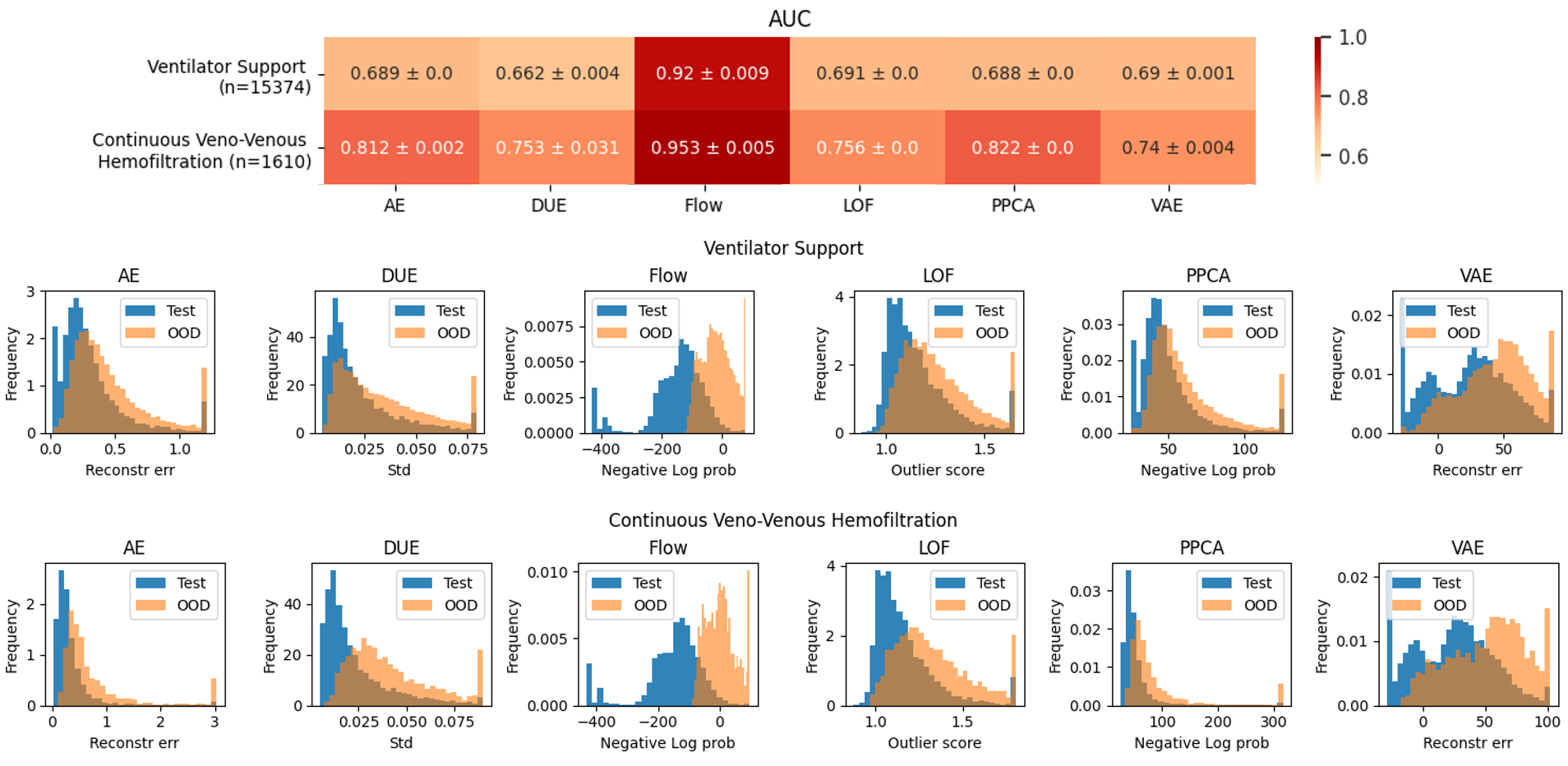}
            \caption{Top panel: Mean AUC scores and standard deviations ($n$=5) of detecting ventilated patients and patients with a renal kidney failure receiving CVVH. Bottom panel: Novelty score distributions. Plotted values are clipped to 5-95\% of test novelty scores to prevent outlier scores from skewing the graphs.}
            \label{fig:ventilated}
        \end{figure*}

        \textbf{Dynamic Inclusion-Exclusion Criteria: Received Treatment}. 
        Next, we selected patients outside dynamic inclusion-exclusion criteria based on the type of received treatment. We compared the OOD detection methods on two categories of treatment-related exclusion criteria: patients connected to a specific device and patients receiving medication interventions.
        
        We first tested whether novelty detectors are able to flag patients connected to a ventilating machine and continuous veno-venous hemofiltration (CVVH) device (Figure \ref{fig:ventilated}). Note that in our experiment, there are no features that would directly indicate whether patients are ventilated or on CVVH, and therefore, OOD detectors must infer this information from the rest of the features. While most models are assigning greater novelty scores to the OOD groups, Flow model has the least overlapping distributions for the groups and in-distribution test data (Figure \ref{fig:ventilated}, bottom panel). We also compared how the detection of the ventilated patients changes in time as they get closer to discharge (Appendix Figure \ref{fig:appendix_vent_days}). 
        
        To test medication-related OOD detection, we evaluated the models in detection patients that fall outside of inclusion-exclusion criteria due to administration of inotropic or vassopressor medication (Appendix Figure \ref{fig:appendix_medic}). Flow flagged these patients with AUC as high as 0.97-0.98\%

        \textbf{Observing Patients with New Diseases}. 
        We use data from the same ICU containing confirmed and suspected COVID-19 patients collected as part of the Dutch Data Warehouse \citep{fleuren_dutch_2021} to test OOD detectors in their ability to flag patients with this disease (Appendix Figure \ref{fig:appendix_covid}). While AE, PPCA, and VAE achieved AUC-ROC above 70\% for COVID-19 patients, Flow was able to achieve over 90\% for both COVID-19 patients and suspects.

\textbf{Inference Time for Real-Time Predictions}. 
        Given that predictions about the novelty of a sample are most useful when given in real-time, we timed OOD detectors in predicting a novelty score for a single sample and when coupled with SHAP (See Appendix Table \ref{table:results_timing}). AE and PPCA are almost 30 times faster in inference and SHAP calculations than the other methods.
        
\section{Discussion}
\label{sec:discussion}
    While it is standard practice to evaluate predictive models on each dataset, there is no established way to compare different OOD detection methods. In this paper, we described a set of practical guidelines on how to create and test OOD detection methods on medical tabular data. 
    
    A similar effort in bridging the gap between research studies and practical implementation was shown for other critical areas such as AI explainability. \cite{mohseni2020multidisciplinary} provided categorization of evaluation strategies of explainable AI, while \cite{gade_explainable_2019} presented practical guidelines for using explainability techniques which were illustrated using different case studies. Other studies have proposed general considerations for deploying machine learning models into clinical practice \cite{chen_how_2019}.

    There are many outstanding questions related to design and selection of OOD detection methods. First, to our knowledge, there is no direct solution to class imbalance of in-distribution data for OOD detection and current methods have to rely on availability of sufficiently large sample size. Second, current models are not well suited for dealing with mixed-type data containing categorical and continuous distributions as they require different likelihood functions. Extending density estimators to mixed-type data \citep{nazabal_handling_2020, ma2020vaem} could greatly enhance OOD detection results. Third, more efficient ways to deal with feature correlation can improve evaluation of interpretability of OOD detectors to prevent feature-importance spread  \citep{lundberg2017unified, aas_explaining_2021}.

    The setup described here assumes some prior knowledge about the possible OOD groups that should be detected as \textit{near}-OOD \citep{winkens2020contrastive}. In addition, we acknowledge that apart from tests described in this paper, other clinically relevant scenarios include detecting samples from different hospitals, and testing for corrupted features (see for example  \cite{ulmer2020trust}).
    
    
    How to aggregate results of different OOD tests and select the most reliable OOD detector across the results is another open-ended and application-specific question. In our experiments on real-life mixed-type EHR data, Flow and VAE performed consistently well across different OOD tests. Given that the performance of density estimators depends on the type of data and dimensionality of the dataset (see Section \ref{sec:methods:recs}), the superior performance of these two models can be explained by a relatively low number of categorical features (less than 10\%), potential non-linear interactions of different features (which could be more problematic for PPCA), and large data size (which can make LOF less efficient). Recently, failures of normalizing flows in assigning higher likelihood to OOD inputs were discussed by several papers \citep{kirichenko_why_2020, zhang_understanding_2021, nalisnick2018deep}. We hypothesize that structured data such as images that give rise to local pixel correlations are more prone to such failures than tabular data. 

    Finally, the guidelines for evaluating OOD detectors for medical data described in this paper can help OOD detectors bridge the gap from theoretical possibility to deployed application, enhancing the safety of AI tools and facilitating the uptake of this new technology in the medical field.
    

\acks{We would like to thank our colleagues at Pacmed and Amsterdam UMC for providing us with the data and insights. We also thank Dennis Ulmer for providing valuable feedback on the manuscript.}

\bibliography{references}

\clearpage

\appendix
\section{Extended Methods}\label{apd:methods}

\subsection{Datasets}
\label{apd:methods:dataset}
     We used a dataset from Amsterdam University Medical Center (AmsterdamUMC). This dataset was chosen to illustrate a real-life medical use case of EHR data, which typically has a lot of features, including categorical, and an unbalanced outcome.
     
     AmsterdamUMC has released a freely accessible de-identified version of the data from 2003-2016. Access to the data and/or the Dutch Data Warehouse against COVID-19 can be requested from Amsterdam Medical Data Science (\href{https://amsterdammedicaldatascience.nl/}{https://amsterdammedicaldatascience.nl/}).
     
     AmsterdamUMC dataset contains retrospective data of patients admitted to Intensive Care Unit (ICU) for each day of their admission stay. This results in $58,142$ rows of data corresponding to $15,753$ patients. For some analyses we sub-select rows of data for patients in their last day of admission at the ICU. We refer to this part of the dataset as AmsterdamUMC-DIS and to the full dataset as AmsterdamUMC-ROW. After eliminating ineligible patients using inclusion-exclusion criteria, AmsterdamUMC-ROW dataset contained $n=40,923$ patients and AmsterdamUMC-DIS contained $n=15,695$ patients.
     
    The total number of features in the dataset is 5269. Features were created by analyzing time series and dividing data into windows of 24 hours. We selected 56 features of which 49 are continuous and 7 categorical. For model training and hyperparameter search, we used a a 0.7, 0.15, 0.15,  split for train, validation, and test split, respectively. All feature names used are shown in the list A1.1 at the end of the document
     
     For the experiment in Figure \ref{fig:appendix_medic}, a second version of AmsterdamUMC dataset was used with $n=18,064$ patients at their last day of ICU admission and $150$ features. This selection includes 84 categorical features and 66 continuous features. We refer to this dataset as AmsterdamUMC-DIS-II. After applying inclusion-exclusion criteria, the dataset contained $n=17,482$ patients.

\subsection{Predictive Task}
    Our main goal is to find the best OOD detectors to flag samples that are not well represented in the training data, and which should not receive a prediction. We define this downstream prediction task to be a binary classification of patients who died or were readmitted back to the ICU within 7 days after discharge. Given that the dataset provides labels of readmission or mortality after discharge for each patient, any predictive model can be trained for this task. The  dataset contains 4.9\% of patients who were readmitted or died at the ICU. Samples of patients at the last day of admission (AmsterdamUMC-DIS) are used for training predictive models. The rationale for this choice is that the predictive model should be used when clinicians are already considering discharge of a patient from the ICU. 
    
\subsection{AUC Score of OOD Detection}
    We use AUC-ROC scores to report the performance of OOD detection methods. First, models are trained on in-distribution training data. Then, they are used to predict novelty scores of in-distribution testing data and each samples in each OOD group. The scores of test in-distribution data and an OOD group are then used to calculate AUC-ROC score by assigning a label 0 to all in-distribution samples and a label of 1 to all samples in an OOD group.

\subsection{OOD Tests for AmsterdamUMC Dataset}
 \textbf{Length of stay}. 
    For the AmsterdamUMC dataset, we created OOD groups based on the admission day at the ICU. In-distribution data for this experiment are eligible patients at their last day at the ICU (AmsterdamUMC-DIS). To create OOD groups, we used AmsterdamUMC-ROW dataset. For each patient we computed the total length of stay at the ICU and bracketed samples based on the number of days before discharge of each patient. Given that patients stayed at the ICU for different number of days, the groups of patients with the highest number of days before discharge contained fewer patients (see patient counts listed in Figure \ref{fig:days_before_discharge}).
    
    \textbf{Discharge destination}. 
     We created four OOD groups by applying the inclusion-exclusion criteria of discharge destination. In-distribution data for this experiment are eligible patients in the AmsterdamUMC-DIS-II dataset. We split excluded patients into four groups: patients transferred to Critical Care Unit (CCU) or ICU of another hospital contained ($n=245$) patients, a group of patients that were transferred to another hospital contained ($n=456$) patients, and two groups of patients that died at the ICU and were sent to a mortuary either with autopsy ($n=4$) or without ($n=6$). 
    
    \textbf{Received treatment: devices}. 
    To split patients on a received device-related treatment, we separated patients based on whether they are connect to a ventilator device ($n=15,374$) or to CVVH ($n=1,610$). In-distribution data for this experiment are eligible patients in the AmsterdamUMC-ROW dataset.
    
    \textbf{Received treatment: medication}. 
     To split patients on a received medication-related treatment we separated patients who were administered Enoximone ($n=57$) or Norepinephrine ($n=774$) in the last 24 hours. In-distribution data for this experiment are eligible patients in the AmsterdamUMC-ROW dataset.

    \textbf{COVID-19 patients and suspects}. 
    To create OOD group with of patients with a new disease on AmsterdamUMC, we used groups of COVID-19 patients ($n=130$) and COVID-19 suspects ($n=140$) which fall outside exclusion criteria. In-distribution data for this experiment are eligible patients in the AmsterdamUMC-ROW dataset.

\subsection{Models}
\label{apd:methods:models}
    Given that discriminator models were shown to underperform on OOD detection for medical tabular data \citep{ulmer2020trust}, we limited our analysis to density estimators. We included models that do not model explicit density function, such as Autoencoder and Local Outlier Factor (LOF).

        \textbf{Autoencoder}. 
        AE was used in combination with a reconstruction error metric. This metric was chosen because it is expected that the model will learn to encode training distribution faithfully whereas samples that are different from training data will not be reconstructed properly and will receive a high reconstruction error. For the AmsterdamUMC dataset, we used the following hyperparameters: both encoder and the encoder had 1 layer with 75 units, the latent space contained 20 dimensions, and the learning rate was 0.007. 
        
        \textbf{Variational Autoencoder}. 
        We implemented a Variational Autoencoder (VAE), one of the most popular density estimation techniques, \citep{kingma2014autoencoding}. Similar to AE, the novelty metric was chosen to be reconstruction error. For experiments on AmsterdamUMC, we used the following hyperparameters: encoder and the encoder contained 3 layers each with 25 units, the latent space contained 10 dimensions, and the learning rate was 0.001. Additionally, we used beta-annealing (a deterministic warm-up; \citeauthor{huan2018improving}, \citeyear{huan2018improving}).
        
        \textbf{Probabilistic Principal Component Analysis}. 
        We used Probabilistic Principal Component Analysis (PPCA; \citeauthor{bishop_1999_ppca}, \citeyear{bishop_1999_ppca}) as a latent variable model with a log likelihood metric. The hyperparameter for PPCA model was only the number of components which was selected as 19 for AmsterdamUMC.

        \textbf{Deep Neural Gaussian Process}. 
        We implemented a deep neural Gaussian Process model called Deterministic Uncertainty Estimation (DUE) according to \citep{vanamersfoort2021improving}. This approach builds on the idea of feature-distance awareness which denotes ability of model to quantify distance of new samples to training data. DUE uses a distance-preserving neural network for feature extraction and models uncertainties using a Gaussian Process in the last layer. Standard deviation was used a novelty metric. The hyperparameters that were used for AmsterdamUMC data were the following: we used the Matern 12 kernel function with 50 inducing points, feature-extracting neural network consisted of 4 layers with 256 units, we set the Lipschitz coefficient at 0.5, and the learning rate at 0.002. 
        
        \textbf{Normalizing Flows}. 
        Normalizing flow models are tractable explicit density estimators that learn training data distribution by series of transformations into a normal distribution. We used Masked Autoregressive Flow as described in \citep{papamakarios2018masked} from an open-source implementation available at \citep{nflows}. The hyperparameters were selected to be the following: 20 layers of composite transformations and reverse permutation, the number of hidden units was 256, we used batch normalization between layers, and the learning rate was set to 0.001.
        
        \textbf{Local Outlier Factor}. 
        LOF \citep{de_vries_finding_2010} does not model density function of training data, it expresses local density of features of a sample and compares this density to closest neighbors. The novelty score is then an average local reachability density of the neighbors denoted as outlier score. The only tunable hyperparameter is the number of closest neighbors which was set to 5.

\section{Extended Results}\label{apd:results}

\begin{figure*}[ht!]
        \centering
        \includegraphics[width=2\columnwidth]{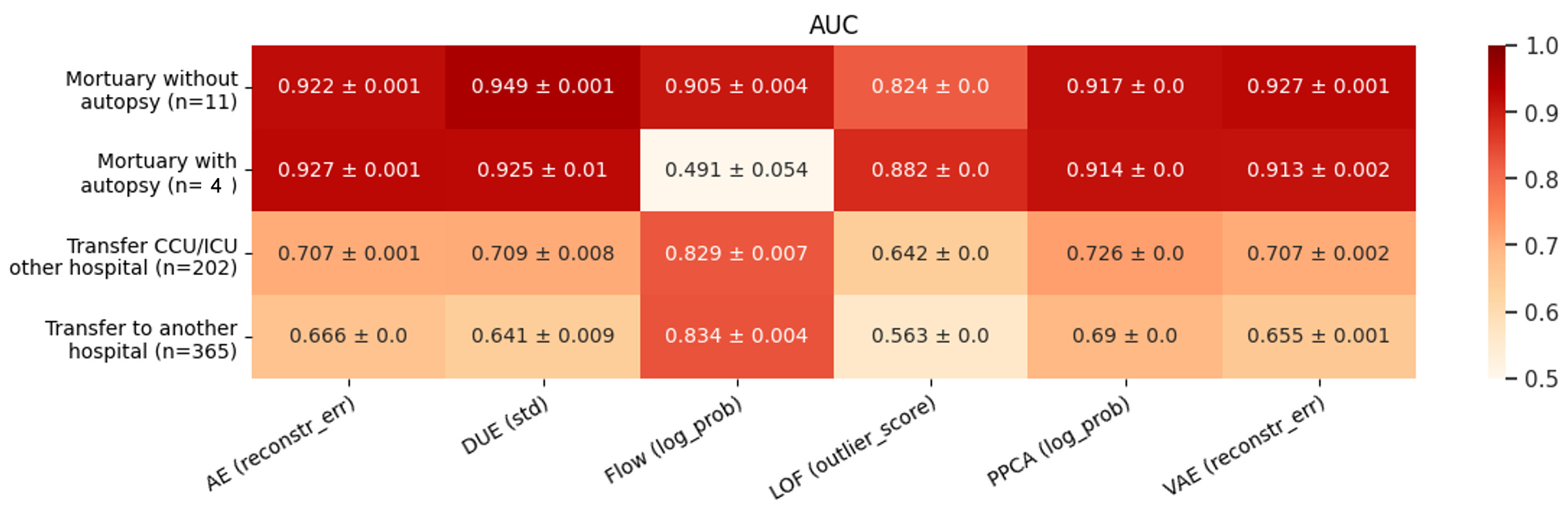}
        \caption{AUC-ROC scores of detecting patients that died at the ICU (Mortuary with or without autopsy) and patients that were transferred to another hospital (ICU or other department). This experiment was performed on the AmsterdamUMC-DIS-II dataset (see dataset description in Appendix \ref{apd:methods:dataset}) which contains more categorical features than AmsterdamUMC-DIS used for the above-described experiments. This could explain the poor performance of Flow in detecting Mortuary with autopsy patients as the model is not suitable for categorical data. }
        \label{fig:appendix_discharge_destin}
    \end{figure*}

\begin{figure*}[t]
        \centering
        \includegraphics[width=2\columnwidth]{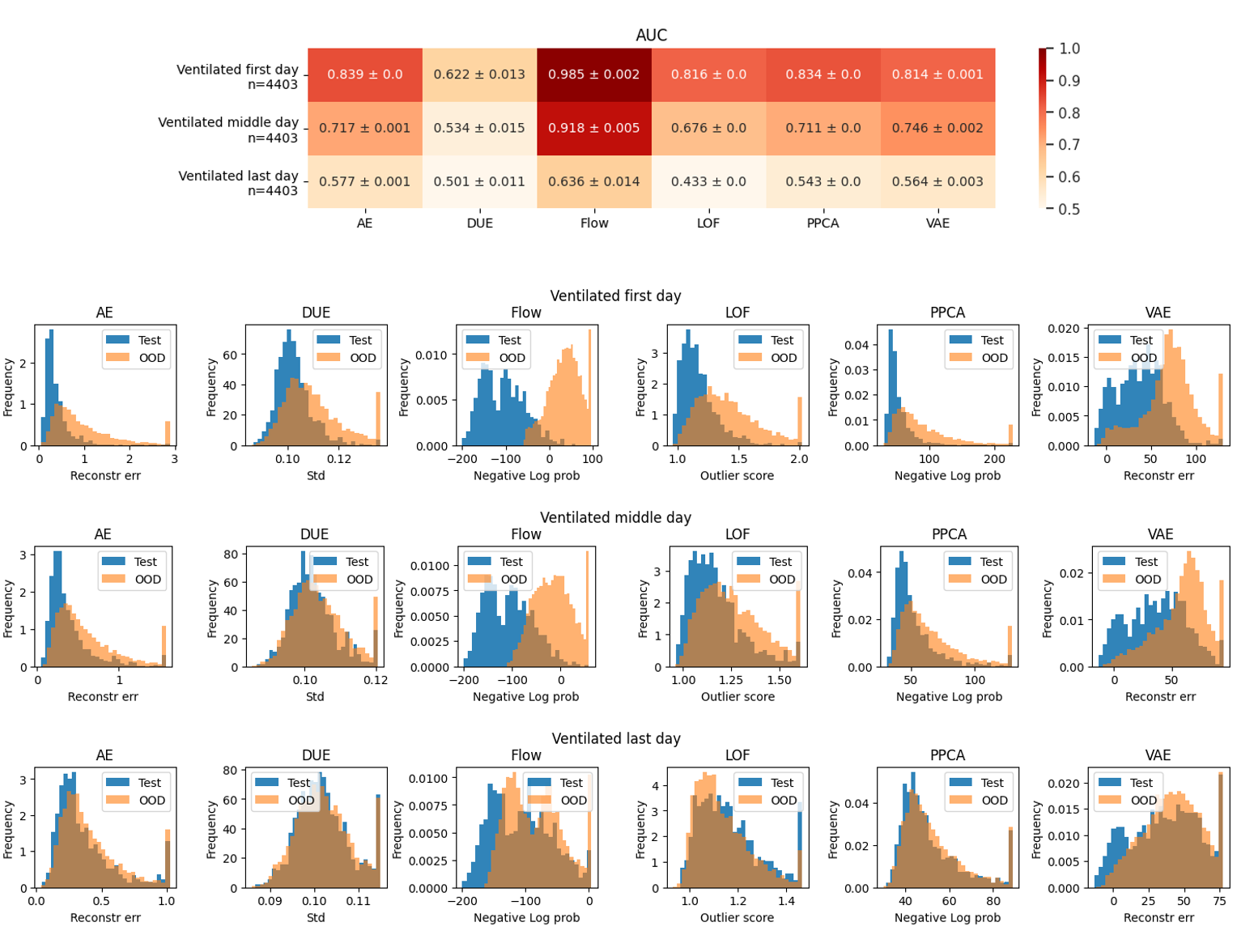}
        \caption{Top panel: AUC-ROC of detecting patients connected to a ventilation machine. \textit{First day} indicates a group of patients in their first day of being connected to the machine, while \textit{last day}  indicates patients that are about to be disconnected from the machine. \textit{Middle day}  is a day in between first and last day of ventilation. Patients with at least 3 days of ventilation were selected. Mean and standard deviation shown ($n$=5). Bottom panel: Averaged novelty score distributions. Plotted values are clipped to 5-95\% of novelty scores to prevent outlier scores from skewing the distributions.}
        \label{fig:appendix_vent_days}
    \end{figure*}

\begin{figure*}[t]
        \centering
        \includegraphics[width=2\columnwidth]{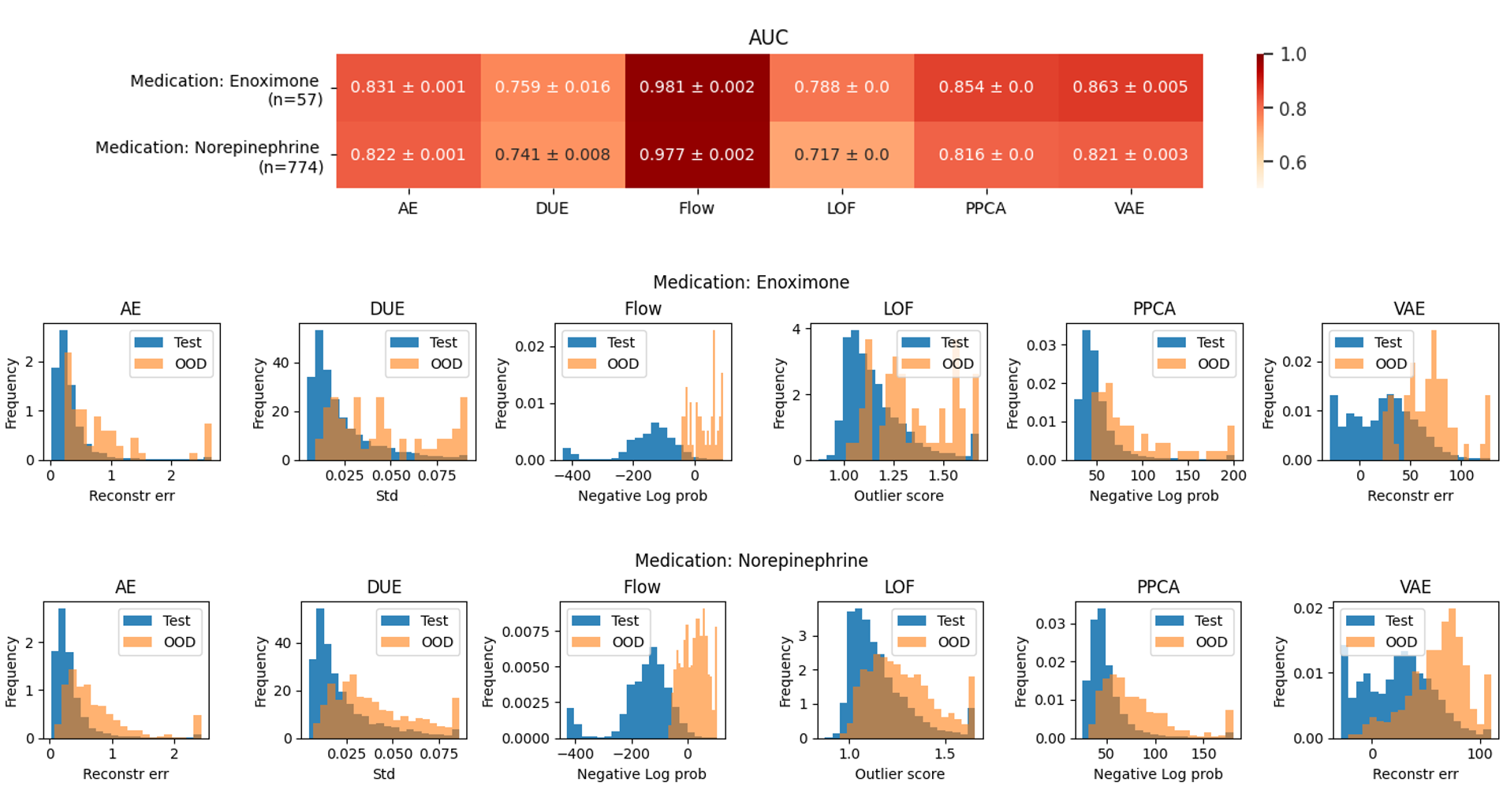}
        \caption{Top panel: AUC-ROC of detecting patients on inotropic or vasopressor medication. Mean and standard deviation shown ($n$=5). Bottom panel: Averaged novelty score distributions. Plotted values are clipped to 5-95\% of novelty scores to prevent outlier scores from skewing the distributions.}
        \label{fig:appendix_medic}
    \end{figure*}

\begin{figure*}[t!]
                \centering
                \includegraphics[width=2\columnwidth]{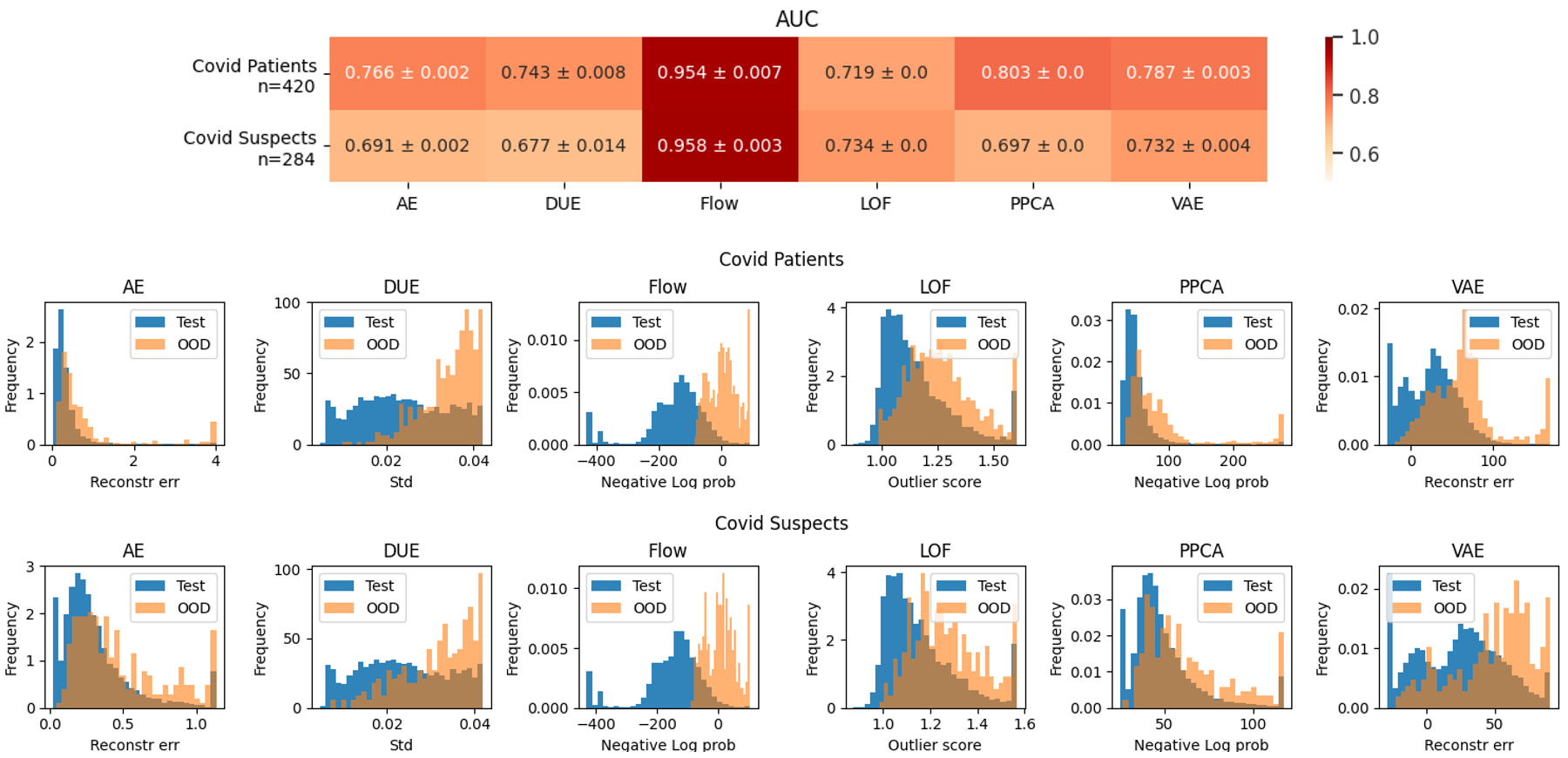}
                \caption{Top panel: AUC scores and standard deviations ($n$=5) of detecting COVID-19 patients and suspects. Bottom panel:  Novelty score distributions. Plotted values are clipped to 5-95\% of test novelty scores.}
                \label{fig:appendix_covid}
    \end{figure*}

    Below, we show figures of the following experiments performed on the AmsterdamUMC which were described in Section \ref{sec:results}:
    
    \begin{itemize}
        \item Dynamic Inclusion-Exclusion Criteria: Discharge Destination (Appendix Figure \ref{fig:appendix_discharge_destin}).
        
        \item Dynamic Inclusion-Exclusion Criteria: Days of Ventilation (Appendix Figure \ref{fig:appendix_vent_days}).
        
        \item Dynamic Inclusion-Exclusion Criteria: Medication 
        (Appendix Figure \ref{fig:appendix_medic}).
    
        \item New disease: COVID-19 patients
        (Appendix Figure \ref{fig:appendix_covid}).
    \end{itemize}

    



    \begin{figure*}[t]
        \centering
        \includegraphics[width=\columnwidth]{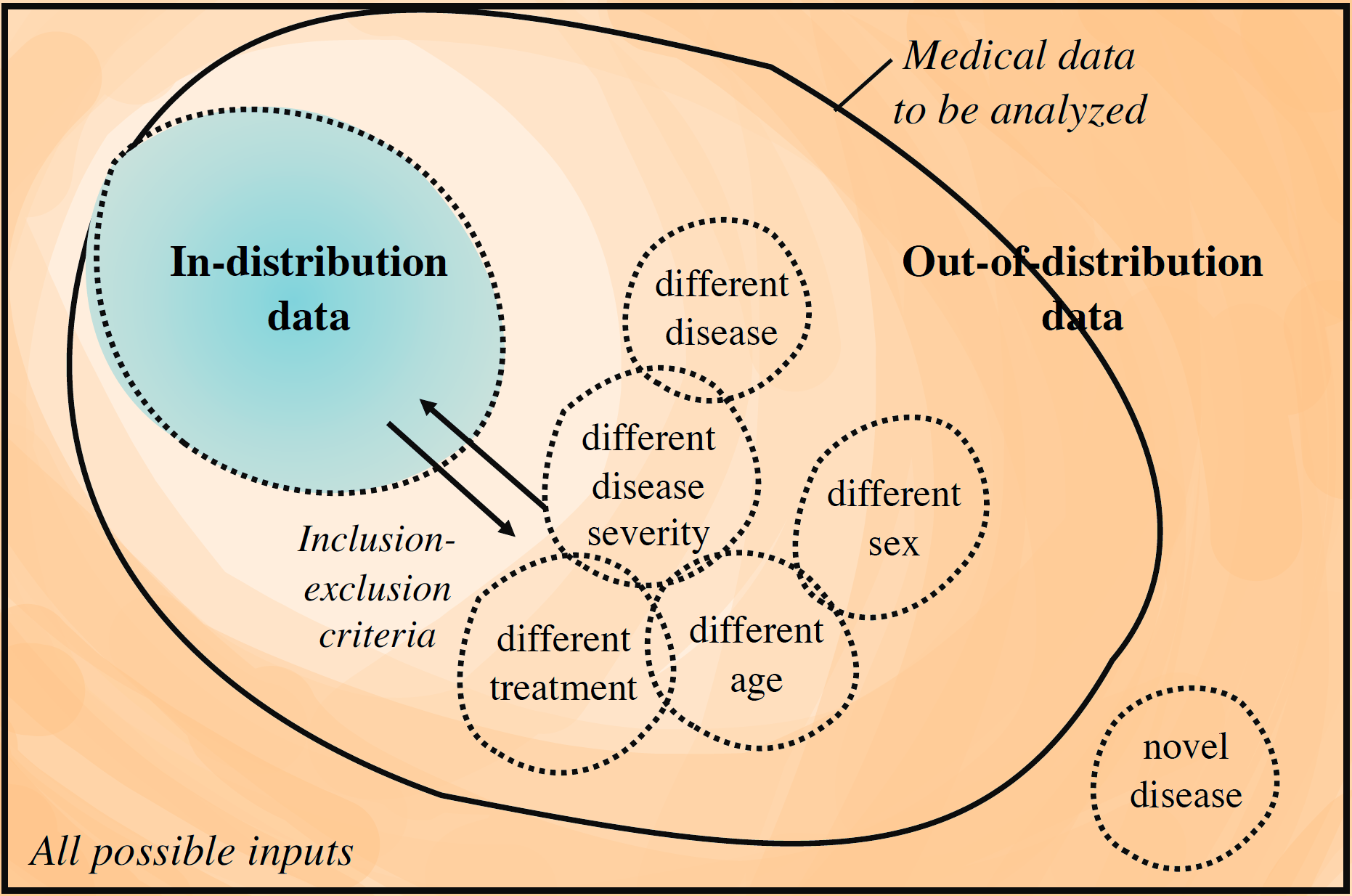}
        \caption{Schematic representation of OOD groups in a medical setting and groups that can be constructed using inclusion-exclusion criteria.}
        \label{fig:scheme}
    \end{figure*}

        \begin{table}[!b]
        \begin{center}
        \caption{Mean and std of user CPU times (in seconds) of inference and SHAP calculation for a single sample.\\}
        \scalebox{0.65}{
        \begin{tabular}{lll}
        \toprule
        {} &        Inference (n=1000) &    SHAP (n=5)\\
        \midrule
        \textbf{AE (reconstr\_err)  } &  0.0006 ± 0.0001 &   1.7583 ± 0.0789 \\
        \textbf{DUE (std)          } &  0.0293 ± 0.0006 &  45.7132 ± 2.4317 \\
        \textbf{Flow (log\_prob)    } &  0.0781 ± 0.0011 &  23.9759 ± 0.7971 \\
        \textbf{LOF (outlier\_score)} &  0.0041 ± 0.0002 &  68.1364 ± 0.2217 \\
        \textbf{PPCA (log\_prob)    } &  0.0016 ± 0.0001 &   1.7503 ± 0.0488 \\
        \textbf{VAE (reconstr\_err) } &  0.1367 ± 0.0005 &  34.1674 ± 1.3905 \\
        \bottomrule
        \end{tabular}
        }
        \label{table:results_timing}
        
        \end{center}
        \end{table}

    \clearpage
    \clearpage
    \textbf{A1.1 AmsterdamUMC Feature List}
    \label{app:methods:feature_list}
         
    \begin{verbatim}
    age
    alanine_transaminase__mean__last_24h
    albumin__mean__last_24h__padded
    alkaline_phosphatase__mean__last_24h
    amylase__mean__last_24h
    arterial_blood_pressure_diastolic__mean__last_24h
    arterial_blood_pressure_systolic__mean__last_24h
    aspartate_transaminase__mean__last_24h
    base_excess__mean__last_24h
    bicarbonate_arterial__mean__last_24h
    bilirubin_total__mean__last_24h
    c_reactive_protein__mean__last_24h
    calcium_ionised__mean__last_24h
    cardiac_output__is_measured__last_24h
    cardiac_output__maximum__last_24h
    cardiac_output__minimum__last_24h
    chloride__mean__last_24h
    cough_stimulus__mode__last_24h__cough_reflex_normal
    cough_stimulus__mode__last_24h__cough_reflex_reasonable
    cough_stimulus__mode__last_24h__cough_reflex_weak
    creatinine__last__overall
    fio2__mean__last_24h
    fluid_out_urine__total_dose__last_24h
    gender_M
    gender_V
    glasgow_coma_scale_total__maximum__last_24h
    glasgow_coma_scale_total__minimum__last_24h
    glucose__mean__last_24h
    heart_rate__mean__last_24h
    hemoglobin__mean__last_24h
    lactate_dehydrogenase__mean__last_24h
    lactate_unspecified__last__overall
    leukocytes__mean__last_24h__padded
    magnesium__mean__last_24h
    neutrophils__mean__last_24h
    o2_flow__mean__last_24h
    o2_saturation__mean__last_24h
    pao2_over_fio2__mean__last_24h
    pco2_arterial__mean__last_24h
    ph_arterial__mean__last_24h
    phosphate__mean__last_24h
    potassium__mean__last_24h
    respiratory_rate_measured__mean__last_24h
    \end{verbatim}

    \clearpage
    \begin{verbatim}
    richmond_agitation_sedation_scale_score__maximum__last_24h
    richmond_agitation_sedation_scale_score__minimum__last_24h
    sodium__mean__last_24h
    temperature_blood__mean__last_24h
    temperature_internal__maximum__last_24h
    temperature_internal__minimum__last_24h
    thrombocytes__mean__last_24h
    tidal_volume__mean__last_24h
    time_since_start
    troponin_t__is_measured__last_24h
    troponin_t__maximum__last_24h
    ureum__mean__last_24h
    ureum_over_creatinine__mean__last_24h
    \end{verbatim}

\end{document}